\pdfoutput=1
\documentclass[11pt]{article}

\usepackage[final]{acl}

\usepackage{times}
\usepackage{latexsym}

\usepackage[T1]{fontenc}
\usepackage[utf8]{inputenc}

\usepackage{microtype}

\usepackage{inconsolata}

\usepackage{graphicx}
\usepackage{amsmath}
\usepackage{graphicx}
\usepackage{placeins}
\usepackage{tcolorbox}
\usepackage{multirow}
\usepackage{enumitem}
\usepackage{subcaption}

\title{OmniFlatten: An End-to-end GPT Model for Seamless Voice Conversation}

\author{Qinglin Zhang, Luyao Cheng, Chong Deng, Qian Chen, Wen Wang, \\ {\bf Siqi Zheng,} {\bf Jiaqing Liu,} {\bf Hai Yu,} {\bf Chaohong Tan,} {\bf Zhihao Du,}  {\bf Shiliang Zhang}  \\ Tongyi Lab \\ {\texttt{\{qinglin.zql, shuli.cly, dengchong.d, tanqing.cq, w.wang\}@alibaba-inc.com}}
}

\begin{document}
\maketitle
\begin{abstract}
Full-duplex spoken dialogue systems significantly surpass traditional turn-based dialogue systems, as they allow simultaneous bidirectional communication, closely mirroring human-human interactions. However, achieving \textit{low latency} and \textit{natural interactions} in full-duplex dialogue systems remains a significant challenge, especially considering human conversation dynamics such as interruptions, backchannels, and overlapping speech. In this paper, we introduce a novel End-to-End GPT-based model \textbf{OmniFlatten} for full-duplex conversation, capable of effectively modeling the complex behaviors inherent to natural conversations with low latency. To achieve full-duplex conversation capabilities, we propose a multi-stage post-training scheme that progressively adapts a text large language model (LLM) backbone into a speech-text dialogue LLM, capable of generating text and speech in real time, without modifying the architecture of the backbone LLM. The training process comprises three stages: modality alignment, half-duplex dialogue learning, and full-duplex dialogue learning. In all training stages, we standardize the data using a flattening operation, which enables unifying the training methods and the GPT backbone across different modalities and tasks. Our approach offers a simple modeling technique and a promising research direction for developing efficient and natural end-to-end full-duplex spoken dialogue systems. Audio samples of dialogues generated by OmniFlatten can be found at this web site~\footnote{\url{https://omniflatten.github.io/}}.
\end{abstract}

\section{Introduction}
\label{sec:intro}

Traditional turn-based spoken dialogue systems only support half-duplex communication, that is, the communication is conducted bidirectionally between user and system (assistant) but \textit{not simultaneously}. These systems, while effective in many real-world applications, often fall short when they come to handle interruptions, backchannels, and overlapping speech, which reflect the spontaneous nature of human-human conversation. Conversely, full-duplex spoken dialogue systems allow \textit{simultaneous} two-way communication, closely mirroring human-human conversations. Full-duplex spoken dialogue systems facilitate \textbf{more natural and efficient communications than turn-based dialogue systems}, by \textit{speaking, listening, and thinking at the same time}. However, achieving \textbf{low latency} and \textbf{natural interactions} in full-duplex dialogue systems remains  a significant challenge.

Recent efforts in developing spoken dialogues systems have been driven by advancements in large language models (LLMs) and can be roughly categorized into \textit{collaborative systems} and \textit{end-to-end (E2E) systems}. Collaborative systems interface LLM-based dialogue modules with external ASR or TTS modules for speech understanding and generation. For example, Qwen-audio ~\cite{chu2024qwen2audiotechnicalreport} takes speech input, outputs text, and converts them to verbal responses via TTS; Mini-Omni2~\cite{xie2024miniomni2opensourcegpt4ovision} investigated a command-based interruption approach to enable full-duplex conversation capabilities.
  In contrast, some E2E systems~\cite{zhang2023speechgptempoweringlargelanguage,xie2024miniomnilanguagemodelshear,fang2024llamaomniseamlessspeechinteraction,zeng2024glm4} directly model speech-to-speech dialogues based on speech-text multimodal models; yet most of these models are turn-based dialogue models and do not support full-duplex conversation. Notable recent progresses in developing E2E full-duplex spoken dialogue systems include SyncLM~\cite{veluri2024turnbasedinterfacessynchronousllms} and  Moshi~\cite{kyutai2024moshi}. Specifically, Moshi models multiple streams of user's speech input and Assistant's text and speech output in parallel, simplifying modeling full-duplex dialogues. However, this parallel framework is not natively supported by GPT models, hence requires sophisticated designs. 
Regarding SyncLM, similar to our approach, it also learns to predict interleaved chunks of user and assistant speech units to acquire real-time full-duplex conversation capabilities. However, 
different from the simple deduplication strategy by SyncLM to improve the model's semantic capabilities, we explore explicit text token prediction to achieve this goal.

\begin{figure*}[h]
    \centering
\begin{minipage}{\textwidth}
            \begin{minipage}{0.48\textwidth}
                \includegraphics[width=\linewidth]{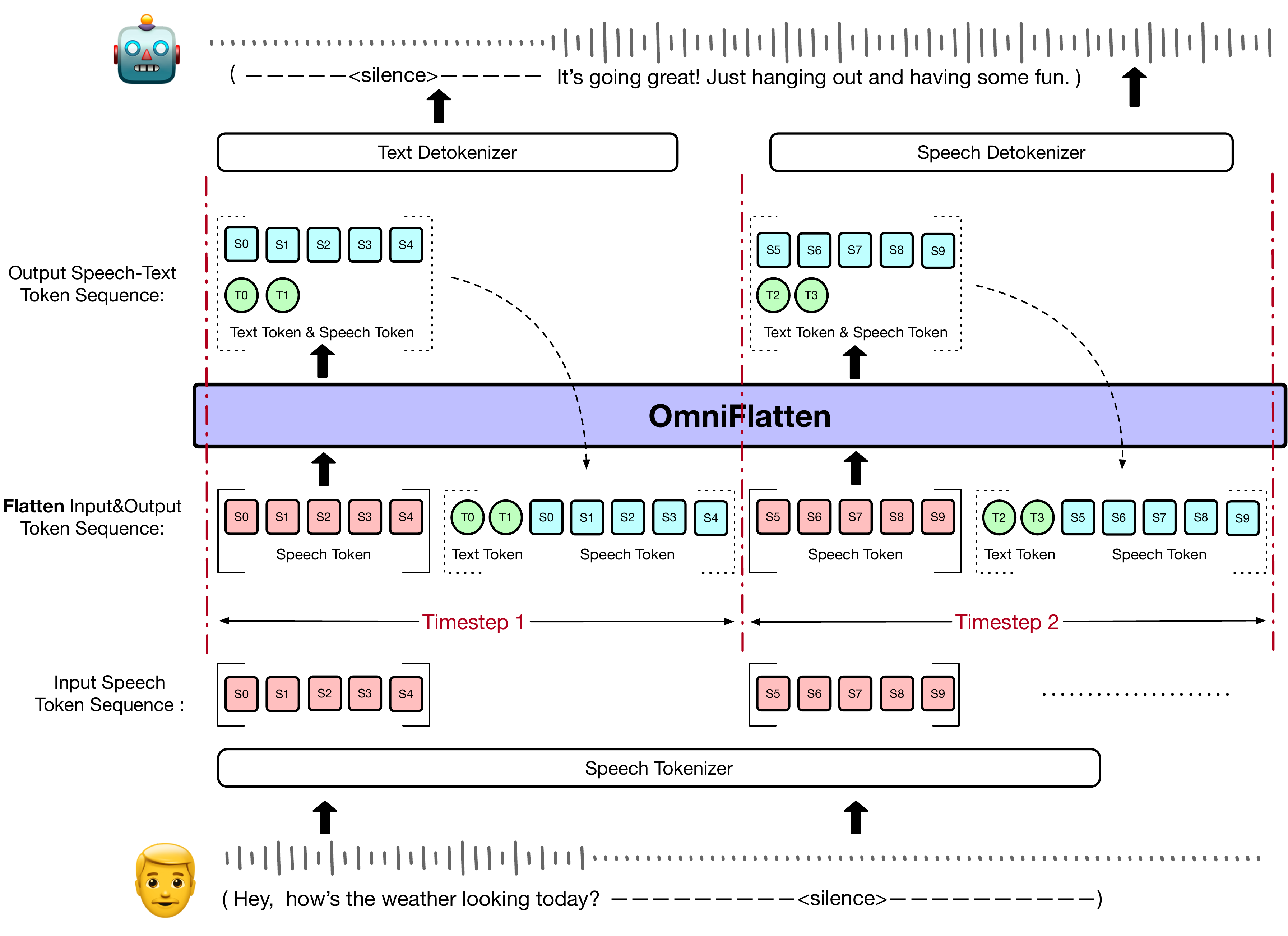}
                \subcaption{The overall architecture of our E2E full-duplex spoken dialogue model \textbf{OmniFlatten}.}
                \label{fig:OmniFlatten}
            \end{minipage}
            \hfill
            \begin{minipage}{0.48\textwidth}
                \includegraphics[width=\linewidth]{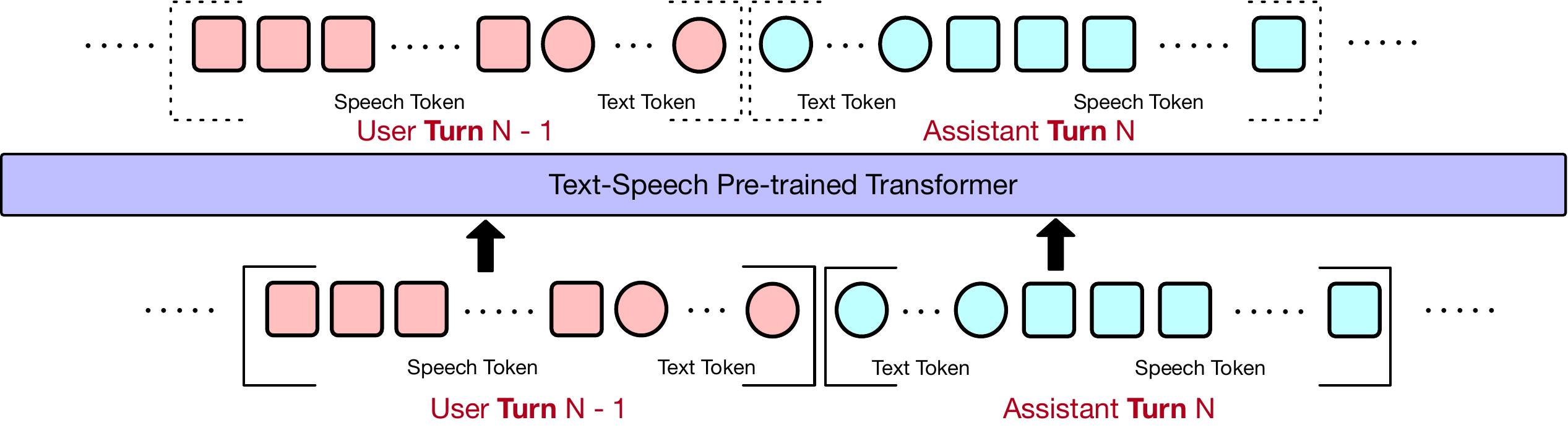}
                \subcaption{Half-duplex Dialogue Training based on all four streams of speech and text tokens of User and Assistant, organized according to the actual \textbf{speaker turns}. We flatten the speech and text tokens into a single sequence, as follows: User Speech Tokens (red squares) and User Text Tokens (red circles) in Turn N-1, Assistant Text Tokens (blue circles) and Assistant Speech Tokens (blue squares) in Turn N.}
                \label{fig:4channel}
            \end{minipage}
        \end{minipage}
        \vspace{5pt}
        % 右侧下排的两张图
        \begin{minipage}{\textwidth}
            \begin{minipage}{0.48\textwidth}
                \includegraphics[width=\linewidth]{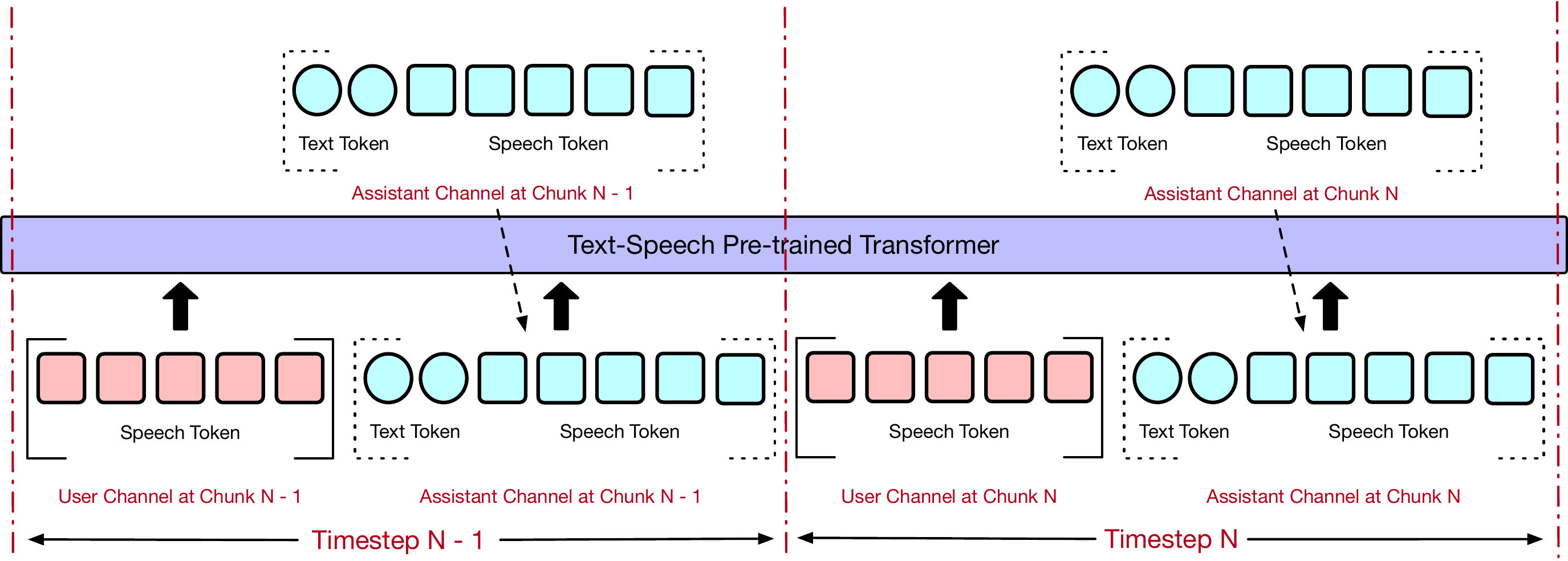}
                \subcaption{Full-duplex Dialogue Training based on \textbf{three streams} of full-duplex dialogue data. User input and Assistant output speech and text token sequences are segmented into short chunks and flattened. At Chunk N-1, five user speech tokens (red squares) are input, and the model outputs two assistant text (blue circles) and five assistant speech tokens (blue squares). The dashed arrows denote that within a chunk, the model appends the predicted Assistant text and speech tokens into input to complete autoregressive decoding.}
                \label{fig:3channel}
            \end{minipage}
            \hfill
            \begin{minipage}{0.48\textwidth}
                \includegraphics[width=\linewidth]{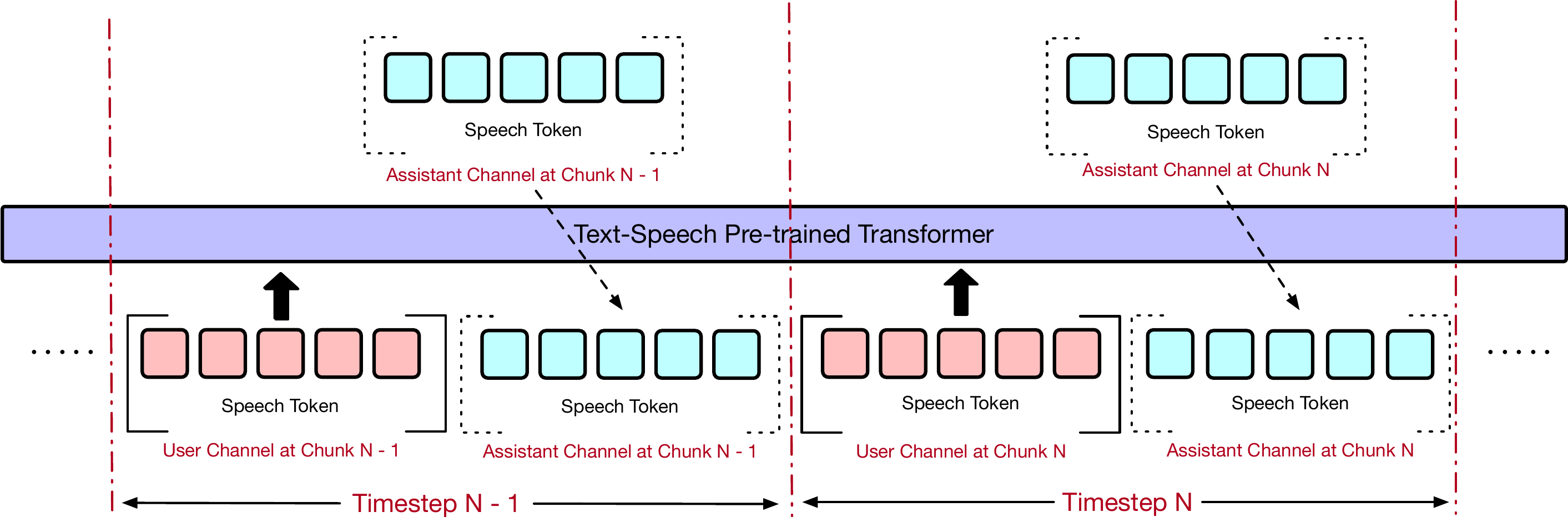}
                \subcaption{Full-duplex Dialogue Training based on \textbf{two streams} of full-duplex dialogue data (further removing the Assistant text stream). In Chunk N-1, five User speech tokens are input, and the model outputs five Assistant speech tokens in Chunk N-1.}
                \label{fig:2channel}
            \end{minipage}
        \end{minipage}
 \caption{The overall architecture of our \textbf{OmniFlatten} and the three dialogue learning stages.}
    \label{figure_model}
\end{figure*}

To address the challenges of achieving \textit{natural interactions} and \textit{low latency} in full-duplex spoken dialogue systems, we introduce a novel E2E GPT-based model \textbf{OmniFlatten} for full-duplex speech conversation. OmniFlatten is capable of effectively learning the complex behaviors inherent to natural conversations and facilitates human-like interactions with low latency. We propose a \textbf{multi-stage progressive post-training scheme} to adapt a text LLM backbone into a robust speech-text dialogue model. The multi-stage post-training process begins with supervised multi-task fine-tuning of the text LLM backbone, using ASR and TTS tasks, to achieve speech-text modality alignment and obtain a multimodal LLM capable of accurately interpreting and generating both speech and text.
After obtaining the speech-text LLM, we fine-tune it using interleaved and flatten dialogues through progressive stages of half-duplex dialogue learning and full-duplex dialogue learning. Notably, throughout all training stages, we standardize dialogue data with this flattening operation, which enables unifying the training methods and the GPT backbone across different modalities and tasks.

Our contributions can be summarized as follows:
\begin{itemize}[leftmargin=*,noitemsep]
\item We propose a novel End-to-End GPT-based model \textbf{OmniFlatten}, capable of effectively modeling the complex behaviors inherent to natural human-like dialogues, with low latency. We propose a \textbf{multi-stage post-training scheme} that successfully adapts a text-based foundation LLM into a robust speech-text dialogue model, by performing supervised multi-task fine-tuning based on ASR and TTS for speech-text modality alignment, then conducting fine-grained chunking of speech and text streams of dialogues and flattening them into a single sequence to progressively train the model to acquire half-duplex and full-duplex conversation capabilities. Notably, \textbf{OmniFlatten does not make any structure modifications to the GPT model, nor relies on computationally intensive pre-training}. Our approach offers a simple modeling technique and a promising research direction for developing efficient and natural E2E full-duplex dialogue systems.
\item We design \textbf{a data synthesis and simulation pipeline to synthesize full-duplex spoken dialogue data} for training and evaluation. 
\item We evaluate the dialogue quality generated by OmniFlatten using high-performing LLMs as evaluators, and evaluate the turn-taking performance, including assistant turn-taking and user turn-taking, as well as the run-time efficiency. The results demonstrate that the dialogues generated by OmniFlatten exhibit reasonable quality, with both modality alignment and half-duplex learning stages improving the model's full-duplex dialogue capabilities. 
\end{itemize}

\begin{comment}
\begin{figure*}[htp]
    \centering
    \includegraphics[width=0.8\textwidth]{gpt-structure-final.pdf}
    % \caption{The overview of \textbf{OmniFlatten} as an end-to-end full-duplex spoken dialogue model. The model constantly receives User input speech and simultaneously outputs Assistant speech and text. The speech tokenizer converts User's speech (including silent segments) into the input speech token sequence (depicted as the red square sequence S0, S1, S2, S3, S4, ...). We input a chunk of input speech tokens (tokens within the solid-line square brackets) into OmniFlatten to decode a chunk of output tokens (tokens within the dashed-line square brackets). The output tokens include both speech tokens (the blue square sequence S0, S1, S2, S3, S4, ...) and text tokens (the green circle sequence T0, T1, T2, ...). We put output text tokens into a fixed text chunk size and then output speech tokens in a fixed speech chunk size, in order to use the predicted text to guide speech generation.  User speech tokens and predicted Assistant text and speech tokens are interleaved according to speaker turns and flatten into a single sequence. Finally, the output speech tokens and text tokens are fed into the speech detokenizer and the text detokenizer, respectively, to obtain Assistant's output speech and text.}
    \caption{\small{The overview of our E2E full-duplex spoken dialogue model \textbf{OmniFlatten}.}}
    \label{fig:OmniFlatten}
\end{figure*}
\end{comment}

\section{Related Work}

\label{sec:related}
In recent years, the field of spoken dialogue models has seen significant advancements driven by the prominent technological advancements in LLMs and particularly speech-text multimodal models. Qwen-audio2~\cite{chu2024qwen2audiotechnicalreport} and SALMONN~\cite{tang2024salmonngenerichearingabilities} support voice input but only output text; hence, they rely on external TTS systems to synthesize speech output. SpeechGPT~\cite{zhang2023speechgptempoweringlargelanguage}, LauraGPT~\cite{du2024lauragptlistenattendunderstand}, Mini-Omni~\cite{xie2024miniomnilanguagemodelshear}, LLaMA-Omni~\cite{fang2024llamaomniseamlessspeechinteraction} and GLM-4-Voice~\cite{zeng2024glm4} are capable of comprehending both speech and text input and generating output in both speech and text; however, they are predominantly turn-based dialogue models and do not support full-duplex conversation. LSLM~\cite{DBLP:journals/corr/abs-2408-02622} explores full-duplex scenarios by integrating TTS models to perform end-to-end modeling of turn-taking task, thereby enabling the model to continuously listen while speaking and stop at any moment.

Recent progresses in developing end-to-end full-duplex spoken dialogue systems include dGSLM~\cite{DBLP:journals/tacl/NguyenKCAHETASM23}, VITA~\cite{fu2024vitaopensourceinteractiveomni}, the open-sourced Moshi~\cite{kyutai2024moshi}, and SyncLM~\cite{veluri2024turnbasedinterfacessynchronousllms}. VITA implements a duplex scheme with two separate modules: one module generates responses to user queries while another module continuously monitors environmental inputs to selectively provide updated interactions. Moshi models multiple streams of user's speech input and system's text and speech output in parallel, allowing for conceptually and practically simple handling of full-duplex dialogues. This approach, compared to its predecessors, offers a more robust solution for managing simultaneous voice inputs and outputs, thereby facilitating more natural and efficient interactions. However, this parallel framework is not supported natively by GPT-based models and hence requires sophisticated designs such as acoustic delay and inner monologue~\cite{kyutai2024moshi}.

Another related concurrent work is SyncLM~\cite{veluri2024turnbasedinterfacessynchronousllms}, which, like our model, employs time-chunking for full-duplex interaction but uses a deduplication strategy for audio sequences. This strategy, while simplifying modeling, leads to errors in audio reconstruction. In contrast, our approach maintains the integrity of discretized audio tokens, thereby preserving audio quality. Additionally, we enhance full-duplex dialogue efficacy through direct modality alignment using ASR and TTS, and progressive learning. Our model concurrently generates both text and audio tokens, unlike SyncLM which only produces speech tokens.

\section{Methodology}
\label{sec:method}

Figure~\ref{fig:OmniFlatten} depicts the overall architecture of our E2E full-duplex dialogue model OmniFlatten. We employ audio tokenizer to discretize each user input and assistant output speech stream of a dialogue into a discrete speech token sequence. We then interleave the speech token sequences with the corresponding text token sequences and flatten them into a single sequence.
Our approach employs a multi-stage progressive training process to adapt a text-based LLM into a robust end-to-end full-duplex spoken dialogue model, through modality alignment and dialogue learning. 

\subsection{Audio Tokenization And Detokenization}
We adopt the speech tokenizer used in CosyVoice\footnote{\url{https://github.com/FunAudioLLM/CosyVoice/tree/main}}~\cite{du2024cosyvoicescalablemultilingualzeroshot, an2024funaudiollmvoiceunderstandinggeneration} since through supervision from multilingual ASR, this speech tokenizer converts speech to semantic tokens, hence benefiting speech understanding and content consistency for speech generation. The tokenizer utilizes an encoder and a Vector Quantization (VQ) layer to discretize audio into speech tokens with a single codebook of 4096 codes. For detokenization, we adopt the same Optimal-transport Conditional Flow Matching model (OT-CFM) used in CosyVoice. OT-CFM transforms the speech token sequence into Mel spectrogram, which is used to generate the final audio output with the HifiGAN vocoder~\cite{DBLP:conf/nips/KongKB20}. Prior works~\cite{DBLP:conf/iclr/LipmanCBNL23, DBLP:journals/tmlr/0001FMHZRWB24} show that OT-CFM outperforms diffusion probabilistic models with simpler gradients, easier training, and faster generation.

\subsection{Modality Alignment}
\label{subsec:modality_alignment}
We start with post-training a pre-trained text LLM backbone to obtain a speech-text multimodal model for speech understanding and speech generation. We use Qwen2-0.5B\footnote{\url{https://huggingface.co/Qwen/Qwen2-0.5B}} as our base model due to its small model size for low compute resource consumption and its competitive performance for models with this small size. We perform supervised fine-tuning (SFT) using paired speech-text data for ASR and TTS tasks. For each speech-text pair $<S_{seq}, T_{seq}>$, we construct ASR training samples as $\quad [ASR][SOS] \text{S\_seq}[EOS][SOT] \text{T\_seq}[EOT]$, and TTS training samples as $\quad [TTS][SOT] \text{T\_seq}[EOT][SOS] \text{S\_seq}[EOS]$,  where [ASR] and [TTS] denote ASR and TTS task IDs; [SOS], [EOS], [SOT], [EOT] are special tokens denoting the Start and the End of the Speech sentence or the Text sentence, respectively.  The speech-text multimodal model is used for the subsequent Dialogue Learning.

\subsection{Dialogue Learning}
\label{subsec:dialogue_learning}

Building upon the speech-text multimodal model, we conduct dialogue learning in three stages, including half-duplex dialogue training using both speech and text streams of turn-based dialogue data, and then full-duplex dialogue training based on fine-grained chunking and alignment of speech and text sequences. Specifically, during full-duplex dialogue training, we first remove the input text stream and use the remaining three streams for training, and then further remove the output text stream and use the remaining two streams for training, in order to gradually eliminate the dependence on text information, focus on speech-to-speech generation.

\subsubsection{Half-duplex Dialogue Training} 
\label{subsubsec:half-duplex-training}

Half-duplex dialogue agents are special and simpler cases of full-duplex dialogue agents, where human and assistant take turns to speak and there is no overlapping speech, that is, during the speaker's turn, the listener is fully silent. Since there is no overlapping speech in the ASR and TTS data used for learning modality alignment, half-duplex dialogue training is more consistent with the aligned multimodal model than full-duplex dialogue training which requires the model to handle turn-taking, backchannel, and overlapping speech. Adopting the concept of curriculum learning, we first conduct half-duplex dialogue training then full-duplex dialogue training. During half-duplex dialogue training, we train the model to essentially perform ASR on the speech tokens of user and obtain the text content, next generate a textual response for assistant based on the text content of user, and then predict the speech tokens for assistant's textual response by basically executing a TTS task. This pattern is extended to multiple turns of a dialogue, as illustrated in Figure~\ref{fig:4channel}.

\subsubsection{Full-duplex Dialogue Training} 
\label{subsubsec:full-duplex-training}

\noindent \textbf{Training on Three-Stream Data} 
%%%%A human-like full-duplex conversational agent is required to handle simultaneous bidirectional conversations with low latency. 
In order to meet the real-time requirements of a full-duplex conversational agent, we remove the user text stream from the four-stream data and train the model with the remaining three steams. In order to handle overlapping speech, we introduce \textit{chunking} and \textit{relaxed speech-text token alignment} based on the chunks; in this way, we do not require strict token-level alignment between speech and text. Specifically, to prepare the training data for this stage, we chunk the speech and text token sequences of dialogues with fixed chunk sizes and then interleave the three-stream data and \textit{flatten} them into a single sequence for training, following the order of \textit{input speech}, \textit{output text}, and \textit{output speech}, as depicted in Figure~\ref{fig:3channel}. The chunking and flattening operation enables our model to stream input speech tokens and output text and speech tokens in real time.  Notably, given the higher efficiency of text, the size of the text chunks is generally smaller than that of the speech chunks. In this work, we set the text chunk size to 2 tokens and the speech chunk size to 10 tokens. This approach ensures that the output text does not excessively precede the speech content, thereby both minimizing the discrepancies with the aforementioned 4-stream data format and maximizing preservation of TTS capabilities. After the end of the text content, we use the special character $silent\_text\_token$ to pad the text stream and use $silent\_speech\_token$ to pad the silent regions of the output speech stream.

\noindent \textbf{Training on Two-Stream Data} 
In order to further reduce latency and eliminate dependence on intermediate text and hence focus on speech-to-speech generation, we further remove the assistant output text stream and retain only the user input and assistant output speech streams. Figure~\ref{fig:2channel} depicts this training process on chunked two-stream data. 

\section{Experiments}

\subsection{Data}
\paragraph{Modality Alignment Dataset} 
The objective of the training stage for modality alignment (Section~\ref{subsec:modality_alignment}) is to help the model learn the correspondence between speech tokens and text tokens and enable the model to acquire two key capabilities: ASR and TTS. To achieve this goal, we combine a set of TTS and ASR datasets that consist of both open-source and proprietary data. The open-source datasets comprise both Mandarin and English data, including Aishell-3~\cite{AISHELL-3_2020}, LibriTTS~\cite{DBLP:conf/interspeech/ZenDCZWJCW19}, TED-LIUM~\cite{DBLP:conf/specom/HernandezNGTE18}, VoxPopuli~\cite{DBLP:conf/acl/WangRLWTHWPD20}, Librispeech~\cite{librispeech}, MLS~\cite{Pratap2020MLSAL} and Wenetspeech~\cite{zhang2022wenetspeech}. Notably, we only use 15\% of the training set of Wenetspeech that has high quality transcripts,  for the ASR task. We incorporate several proprietary ASR and TTS datasets. In sum, the datasets for the speech-text modality alignment include about 100K hours of audio. Among all data, approximate 30\% are open-source data and 70\% are proprietary data.

\paragraph{Simulated Voice Chat Dataset}
\begin{figure*}
    \centering
    \includegraphics[width=1\linewidth]{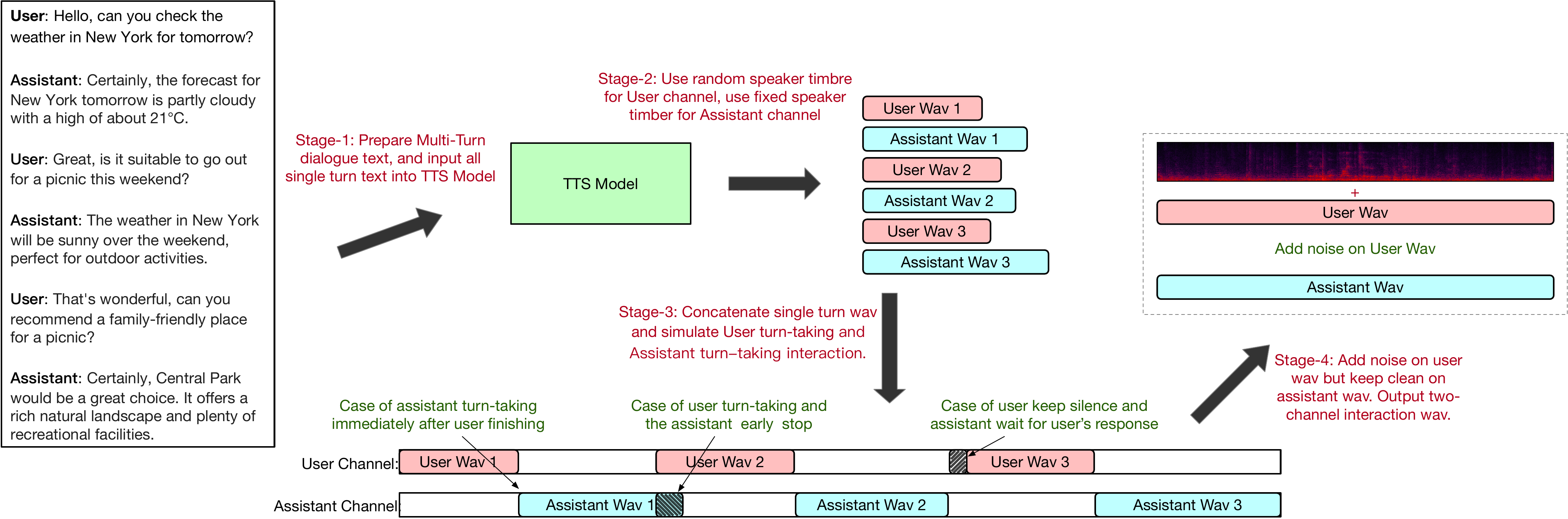}
    % \caption{The overview of \textbf{OmniFlatten} as an end-to-end full-duplex spoken dialogue model. The model constantly receives User input speech and simultaneously outputs Assistant speech and text. The speech tokenizer converts User's speech (including silent segments) into the input speech token sequence (depicted as the red square sequence S0, S1, S2, S3, S4, ...). We input a chunk of input speech tokens (tokens within the solid-line square brackets) into OmniFlatten to decode a chunk of output tokens (tokens within the dashed-line square brackets). The output tokens include both speech tokens (the blue square sequence S0, S1, S2, S3, S4, ...) and text tokens (the green circle sequence T0, T1, T2, ...). We put output text tokens into a fixed text chunk size and then output speech tokens in a fixed speech chunk size, in order to use the predicted text to guide speech generation.  User speech tokens and predicted Assistant text and speech tokens are interleaved according to speaker turns and flatten into a single sequence. Finally, the output speech tokens and text tokens are fed into the speech detokenizer and the text detokenizer, respectively, to obtain Assistant's output speech and text.}
    \caption{The simulation process of dialogue learning data.}
    \label{fig:simulation_process}
\end{figure*}
%%%%\paragraph{Real-Time Chat Dataset}
For constructing the voice chat data for Dialogue Learning (Section~\ref{subsec:dialogue_learning}), we design a data synthesis and simulation pipeline to synthesize dialogue data, as shown in Figure~\ref{fig:simulation_process}.
%%%%Regarding the real-time Chat task, we designed a synthetic and simulated pipeline to obtain data for experiments.
Firstly, we collect a substantial amount of high-quality open-source textual dialogue data for subsequent speech synthesis, including Alpaca~\cite{DBLP:journals/corr/abs-2304-03277}, Moss~\cite{MIR-2023-12-294}, BelleCN~\cite{ji2023better}, and ultraChat~\cite{DBLP:conf/emnlp/DingCXQHL0Z23}. We then use heuristic rules and filter out samples that are inappropriate for TTS, for example, samples that comprise a high percentage of non-text elements such as code and mathematical expressions, samples with more than 200 words in English or Chinese, and also samples containing rare or unusual symbols. In the end, we retain approximately 390K multi-turn sessions of turn-based dialogues (half-duplex dialogues).

% Secondly, we input multi-turn dialogue texts into a TTS system to generate single-turn speech. The TTS system utilized in this process is CosyVoice.
Secondly, we use CosyVoice to synthesize speech from turn-based dialogue texts. Regarding timbre selection, the voice timbre for the user channel was sampled from Librispeech~\cite{librispeech} and 3D-Speaker~\cite{zheng20233dspeaker} datasets, whereas a fixed timbre was employed for the assistant channel.

Thirdly, upon obtaining the audio for each turn, we designed a simulation approach to place the audio at specific times, thereby emulating the user-assistant interaction dynamics typically seen in real human-machine interactions. This method encompasses several key situations: (1) The user finishes asking a question, followed by the assistant's immediate response. (2) The user tries to interrupt, leading the assistant to stop speaking abruptly. (3) Once the assistant has finished speaking, it maintains silence while waiting for the user's speaking.

Finally, to mimic real-world scenarios of the user audio channel, we also sample background noise from the MUSAN noise dataset~\cite{musan2015} and add noise into the user audio channel with the signal-to-noise ratio (SNR) between 15 dB and 30 dB. 
Based on this data synthesis and simulation pipeline, we generate a total of \textbf{2,000 hours of multi-channel spoken dialogue data}. We randomly select 1\% of this dataset as the validation set and another 1\% as the test set, and the remaining data are used as the training set.

\subsection{Training and Inference Setup}
We use Qwen2-0.5B~\cite{yang2024qwen2technicalreport} as the base model. During the modality alignment training phase, the maximum sequence length is set to 1024 tokens. In the dialogue learning phase, the maximum sequence length is extended to 8192 tokens. We use the standard cross-entropy loss as our training objective in all stages. Additionally, during the dialogue learning phase, we apply loss masking on the user channel, as we observe that this operation enhances the stability of model training, probably due to the presence of noisy audio input in the user channel. We employ the AdamW optimizer with a weight decay of 0.1, $\beta_1$ of 0.9, and $\beta_2$ of 0.95. The maximum learning rate is set to 2e-05, with warm-up and cosine-decay. We train the model with 5 epochs, with the best model selected based on loss on the validation set. Each batch contains 100M tokens.

During inference, to obtain the assistant textual response prediction from the model, we use the ground truth user channel speech in the test set as the fixed speech input and alternately fill in the predicted assistant speech and text with the fixed speech chunk and text chunk sizes.

\begin{table}[]
\centering
\caption{ASR results on Librispeech and Wenetspeech test datasets. For Librispeech, we report the WER metric. For Wenetspeech, we report the CER metric.}
\label{tab:asr_results}
\resizebox{\columnwidth}{!}{%
\begin{tabular}{ccccc}
\hline
\multirow{2}{*}{Model} & \multicolumn{2}{c}{Librispeech} & \multicolumn{2}{c}{Wenetspeech} \\ \cline{2-5} 
                       & test-clean     & test-other     & test-meeting     & test-net     \\ \hline
Whisper-S              & 3.13           & 7.37           & 25.62            & 16.66        \\
Whisper-L              & 1.82           & 3.5            & 18.87            & 10.48        \\
VITA                   & 8.14           & 18.4           & 12.15            & 16.53        \\
OmniFlatten            & 7.91           & 19.21          & 26.1             & 19.0         \\ \hline
\end{tabular}%
}
\end{table}

\begin{table}[]
\centering
\caption{TTS results on LibriTTS and AIShell-3. For LibriTTS, we utilize whisper-large-v3 model to perform recognition on TTS outputs and assess the WER metric. For AIShell-3, we deploy the paraformer-zh model to recognize TTS results and evaluate CER metric.}
\label{tab:tts_results}
\resizebox{\columnwidth}{!}{%
\begin{tabular}{ccc}
\hline
Model       & LibriTTS(WER) & AIShell-3(CER) \\ \hline
Original    & 2.66                           & 2.52                       \\
ChatTTS     & 8.32                           & 3.87                       \\
CosyVoice   & 2.89                           & 3.82                       \\
OmniFlatten & 4.51                           & 4.46                       \\ \hline
\end{tabular}%
}
\end{table}

\subsection{Evaluations}
\paragraph{Evaluation of ASR and TTS Performance after Modality Alignment.} The Modality Alignment training stage (Section~\ref{subsec:modality_alignment}) aims to help the model learn the correspondence between speech tokens and text tokens and acquire ASR and TTS capabilities; hence, we evaluate the effectiveness of this training stage by evaluating the ASR and TTS performance of the multimodal-aligned model.

For TTS evaluation, we use the model to generate speech tokens based on the input text, which are then synthesized into audio. We follow the experimental setup outlined in the CosyVoice paper: the synthesized audio is subsequently recognized using the Whisper-Large-V3 model~\cite{DBLP:conf/icml/RadfordKXBMS23}~\footnote{\url{https://huggingface.co/openai/whisper-large-v3}\label{whispersharedfootnote}} and Paraformer-zh model~\cite{DBLP:conf/interspeech/GaoLWLSCLZDZ23}~\footnote{\url{https://huggingface.co/funasr/paraformer-zh}}, for English and Chinese datasets, respectively, and the resulting ASR transcripts are scored against the input text to compute Word Error Rate (WER) for English datasets and Character Error Rate (CER) for Chinese datasets. WER/CER could measure the synthesis accuracy and robustness of a model's TTS capabilities, and also reflect the audio quality to a significant extent. Since the main goal of this work is learning conversation dynamics of full-duplex voice chat, we have not adopted standard evaluation metrics for evaluating TTS speech quality, such as Mean Opinion Score (MOS).

For ASR evaluation, we discretize the input speech into discrete speech tokens, then use the model to decode the speech tokens into text. We compare the speech-text aligned multimodal model after Modality Alignment (denoted by \textbf{OmniFlatten}) with Whisper-small~\footnote{\url{https://huggingface.co/openai/whisper-small}} and Whisper-Large-V3 models.  In addition, we have conducted a comparison of ASR capabilities with those reported by VITA. For TTS evaluation, We conduct comparisons between our approach and two competitive TTS systems, namely, ChatTTS\footnote{\url{https://github.com/2noise/ChatTTS}} and CosyVoice. As shown in Table~\ref{tab:asr_results} and~\ref{tab:tts_results}, OmniFlatten demonstrates considerable performance in both ASR and TTS tasks. These results indicate that the Modality Alignment training stage effectively transitions the unimodality text-based LLM to speech-text multimodal model with reasonable speech comprehension and generation capabilities for the following dialogue learning.

\paragraph{Evaluation of Full-duplex Conversation Capability after Modality Alignment, Half-duplex Dialogue Learning, and Full-duplex Dialogue Learning.}
As described in Section~\ref{subsubsec:full-duplex-training}, the training stage of full-duplex dialogue learning on three-stream data helps the model acquire full-duplex dialogue capacity and this model generates both speech and text. Prior works~\cite{DBLP:conf/nips/ZhengC00WZL0LXZ23} demonstrate that competitive text-based LLMs can be reliable evaluators for various natural language generation tasks, as scores assigned by LLM evaluators on generated text show high correlations with human evaluations. Therefore, we evaluate the full-duplex dialogue capabilities of OmniFlatten, by prompting a competitive text LLM to evaluate the \textbf{semantics of dialogues} and assign a score on the predicted assistant response from the model after training on three-stream data. 
We also compared three models: LLaMA-Omni, Moshi, and GLM-Voice in both Chinese and English chat capabilities. To investigate the effects of adding audio modality on the chat capability of models, we examined the text-only chat performance of Qwen2-0.5B-Instruct and Qwen2-7B-Instruct.

The scoring mechanism involves designing a specific prompt and utilizing a competitive text LLM, QWen-max model~\footnote{\url{https://help.aliyun.com/zh/model-studio/developer-reference/use-qwen-by-calling-api}}, to rate the responses from the model on a scale of 1 to 10 points. The specific prompt we use for LLM scoring is detailed in Appendix~\ref{appendix_auto_evaluation_prompt}. We carefully design the prompt to evaluate \textit{fluency} and \textit{coherence} of the predicted assistant response. Given that the full-duplex model may be disrupted by user speech in multi-turn dialogues, which could affect the assessment of chat capabilities, we exclusively focus on assessing the single-turn chat performance of the model. Given the dual output of speech and text by the model, our evaluation encompasses scoring the text output as well as converting the speech output into text using an ASR model specific to the output language, similar to the methodology described in the TTS evaluation section. The outcomes are reported separately as LLM Score (Text) and LLM Score (Speech).

% We also report the CE loss of the model on the test set.  

To analyze the impact of the modality alignment training stage (Section~\ref{subsec:modality_alignment}) and the half-duplex dialogue learning stage (Section~\ref{subsec:dialogue_learning}) on the full-duplex conversation capability of OmniFlatten,  we compare the LLM scores on the predicted assistant responses from the following models: (1) Trained directly on three-stream data (\textbf{OmniFlatten directly 3-stream}). (2) Trained with modality alignment and full-duplex three-stream dialogue data (\textbf{OmniFlatten 3-stream w/o half-duplex}). (3)Trained with modality alignment, half-duplex dialogue data, and full-duplex three-stream dialogue data (\textbf{OmniFlatten 3-stream full process}). (4) Trained with modality alignment, half-duplex dialogue data, and full-duplex three-stream dialogue data and then on two-stream data (\textbf{OmniFlatten 2-stream full process}).
    % \item Ground truth textual response in the test set (denoted by \textbf{GT Response}).

\begin{table*}[]
\centering
\resizebox{\textwidth}{!}{%
\begin{tabular}{lccccc}
\hline
\multirow{2}{*}{Model} & \multirow{2}{*}{Params} & \multicolumn{2}{c}{En}              & \multicolumn{2}{c}{Zh}              \\ \cline{3-6} 
                       &                         & Score(Text) & Score(Speech + ASR) & Score(Text) & Score(Speech + ASR) \\ \hline
Qwen2-0.5B-Instruct    & 0.5B                    &      6.75       &         -         &       6.98      &         -         \\
Qwen2-7B-Instruct    & 7B                      &      8.37       &         -         &       8.09      &         -         \\ 
LLaMA-Omni             & 8B                      &      6.01       &        5.50       &       4.17      &         3.89      \\
Moshi                  & 7B                      &      3.92       &        3.46       &       -         &         -         \\
GLM-Voice            &  9B                     &      6.97       &        6.40       &       7.02      &         6.69      \\ \hline
OmniFlatten directly 3-stream & 0.5B            &       2.99          &      2.59              &        4.94      &         3.95       \\
OmniFlatten 3-stream w/o half-duplex & 0.5B   &      3.89       &        3.54       &       5.25      &       4.76         \\
OmniFlatten 3-stream full process    & 0.5B  &      4.88       &        3.92       &        5.6      &         5.15      \\
OmniFlatten 2-stream full process    & 0.5B                &      -          &     2.19              &       -         &       3.06           \\ \hline
Ground Truth Response            & -                       &      7.65       &          -        &        6.83     &          -        \\ \hline
\end{tabular}%
}
\caption{Performance results of speech and text chat capabilities in both Chinese and English test datasets.}
\label{tab:chat_capacity}
\end{table*}

\begin{table*}[]
\centering
\resizebox{\textwidth}{!}{%
\begin{tabular}{lcccc}
\hline
Models      & \begin{tabular}[c]{@{}c@{}}Assistant Turn-taking Acc@K\\ 1/5/10/25 (\%)\end{tabular} & \begin{tabular}[c]{@{}c@{}}Average Assistant Turn-taking\\ Response Time (ms)\end{tabular} & \begin{tabular}[c]{@{}c@{}}User Turn-taking Acc @K\\ 1/5/10/25 (\%)\end{tabular} & \begin{tabular}[c]{@{}c@{}}Average User Turn-taking\\ Response Time (ms)\end{tabular} \\ \hline
Moshi       & 2.9/18.8/38.5/55.1                                                                   & 553                                                                                          & 0.0/6.2/14.8/45.7                                                                & 753                                                                                     \\
OmniFlatten & 20.6/53.6/ 66.3/71.7                                                                 & 193                                                                                          & 10.9/30.9/41.8/51.8                                                              & 287                                                                                     \\ \hline
\end{tabular}%
}
\caption{Assistant Turn-taking and User Turn-taking accuracy at the k-th token (Acc@K) and Response Time.}
\label{tab:response_capacity}
\end{table*}

The results, as displayed in Table~\ref{tab:chat_capacity}, reveal that the multi-stage training approach enhances performance when comparing several 3-streaming models with OmniFlatten. Specifically, the OmniFlatten directly 3-stream model, lacking speech-text alignment, scores lowest; the absence of a half-duplex stage further diminishes its chat capabilities. The 2-stream experiments show a significant performance drop when models generate only speech outputs, with these models occasionally incorporating unclear semantic content in speech endings, adversely affecting scores. When comparing models that produce both speech and text, the performance in the speech modality consistently falls short of that in the text modality, indicating a slight compromise in accuracy despite the ability to synchronize speech with text outputs.

OmniFlatten exhibits superior chat performance on the English test set compared to Moshi, which frequently resorts to counter-questioning the user instead of providing direct responses. In contrast, compared with the 7B parameter LLaMA-Omni, OmniFlatten lagging scores on the English dataset while leading in the Chinese dataset. The proficiency in Chinese for LLaMA-Omni predominantly stems from the speech-text alignment process, despite the absence of Chinese data in its dialogue learning process. Comparatively, against the 9B parameter GLM-Voice model, OmniFlatten's indicators are relatively underperforming, a situation we attribute to potential data leakage during GLM-Voice's training process involving our test set. Additionally, comparing with the Qwen2-0.5B-Instruct results, the inclusion of the speech modality results in a noticeable decline in chat capabilities.

\paragraph{Turn-taking Performance and Runtime Efficiency.}

To assess the naturalness of OmniFlatten's full-duplex interactions, we evaluate the assistant's ability to respond promptly after the user finishes speaking and to cease speaking when the user attempts to interrupt. In real-world scenarios, timely responses by machines prompt specific human interactive behaviors. For instance, if a machine fails to answer a question quickly, a human might repeat the question. Likewise, if a machine does not stop talking when interrupted, the human may try to interrupt again. Our current testing approach involves using a predefined user speech input while setting a 1.5-second threshold. We consider it a failure of user turn-taking or assistant turn-taking if, exceeded by either, the machine fails to respond (i.e., start or stop answering) within the threshold. Based on this principle, we define the following metrics:

\textbf{Assistant Turn-taking Acc@k:} This metric is defined as whether assistant correctly predicts a non-silence token at the k-th token after the end of a semantically meaningful speech token from user, indicating that assistant starts speaking.

\textbf{User Turn-taking Acc@k:} This metric is defined as whether assistant correctly outputs a silence token at the k-th token after a semantically meaningful speech token is input from user while assistant is speaking. This metric indicates that assistant has successfully responded to the turn-taking attempt from user by stopping assistant's speech and listening.

The evaluation results are shown in Table~\ref{tab:response_capacity}. We compared the response times of OmniFlatten and Moshi during interactions involving user turn-taking and assistant turn-taking scenarios. It was observed that OmniFlatten demonstrates a shorter average response time compared to Moshi in both user and assistant turn-taking contexts. However, in terms of the user turn-taking metric, although OmniFlatten leads in response time and accuracy, neither model exhibits a particularly high success rate in achieving user turn-taking within 25 tokens.

\section{Conclusion}

In this paper, we introduce an end-to-end full-duplex spoken dialogue model OmniFlatten based on synthesizing full-duplex spoken dialogue data and designing a multi-stage progressive training paradigm for modality alignment and dialogue learning. Our approach offers a straightforward full-duplex modeling scheme as it does not change the architecture of the backbone text-based LLM nor relies on computationally intensive pre-training. Empirical evaluations show that the proposed approach is promising for developing end-to-end models to handle full-duplex interactions. In future work, we plan to refine the data synthesis pipeline and better simulate the complex interaction patterns, such as user backchannels. Additionally, we will explore more potentials of this modeling scheme by extending to full-duplex interaction involving more modalities, such as vision.

\section{Limitations}
We acknowledge the limitations of this paper. Our work is based on training with the qwen2-0.5B model and utilizes a relatively small scale of dialogue training data. Compared to other similar models that have been scaled up, our model exhibits considerable potential for enhancement in chat capabilities. Particularly, the response speed of our model, especially in scenarios involving user turn-taking, requires further optimization. We have identified certain conflicts between the goal of early stopping by the assistant and some training objectives in speech-text alignment. Moreover, our current model is not yet capable of handling more complex human-machine interaction phenomena such as user back-channel or even assistant back-channel, indicating that the naturalness of human-machine interaction could be further enhanced.

\bibliography{custom}
\bibliographystyle{acl_natbib}

\appendix
\label{sec:appendix}

\section{Prompt LLM for Dialogue Quality Evaluation}
\label{appendix_auto_evaluation_prompt}

\begin{tcolorbox}[colback=blue!5!white, colframe=blue!75!black, title=Auto-Evaluation Prompt]
Please rate the given dialogue context and response from the voice dialogue system based on the following criteria (1-10 points), and provide a brief evaluation:
\\\\
1. Relevance: Is the response relevant to the query? Is the content related?

2. Accuracy: Does the response correctly address the user's query and provide accurate information?

3. Completeness: Does the response comprehensively cover all aspects of the query?

4. Conversational Nature: Is the response easy to understand, concise, clear, and fluent?
\\\\
Context: {context}

Response: {response}
\\\\
Output in JSON format:

\{\\
  "Strengths": "Positive aspects of the response",\\
  "Weaknesses": "Negative aspects of the response",\\
  "Overall Evaluation": "Overall assessment of the response",\\
  "Total Score (out of 10, directly provide the score)": ""\\
\}

\end{tcolorbox}

\end{document}